\documentclass[final,5p,times,twocolumn,authoryear]{elsarticle}

\usepackage{amssymb}
\usepackage{color,array}

\usepackage{graphicx}

\usepackage{amsmath}
\usepackage{float}
\usepackage{multirow}

\usepackage[export]{adjustbox}
\usepackage{longtable}
\usepackage{subfigure}
\usepackage{placeins}

\journal{Machine Learning with Applications}    
\begin{document}

\begin{frontmatter}



\title{Machine learning for sports betting: should model selection be based on accuracy or calibration?}


\author{Conor Walsh\corref{cor1}\fnref{label1}}
\ead{conorwalsh206@gmail.com}
\author{Alok Joshi\fnref{label1}} 
\ead{aj2151@bath.ac.uk}

\affiliation[label1]{organization={Department of Computer Science, University of Bath},%Department and Organization
            state={Somerset},
            country={UK}}

\begin{abstract}
Sports betting's recent federal legalisation in the USA coincides with the golden age of machine learning. If bettors can leverage data to reliably predict the probability of an outcome, they can recognise when the bookmaker's odds are in their favour. As sports betting is a multi-billion dollar industry in the USA alone, identifying such opportunities could be extremely lucrative. Many researchers have applied machine learning to the sports outcome prediction problem, generally using accuracy to evaluate the performance of predictive models. We hypothesise that for the sports betting problem, model calibration is more important than accuracy. To test this hypothesis, we train models on NBA data over several seasons and run betting experiments on a single season, using published odds. We show that using calibration, rather than accuracy, as the basis for model selection leads to greater returns, on average (return on investment of +34.69\% versus -35.17\%) and in the best case (+36.93\% versus +5.56\%). These findings suggest that for sports betting (or any probabilistic decision-making problem), calibration is a more important metric than accuracy. Sports bettors who wish to increase profits should therefore select their predictive model based on calibration, rather than accuracy.

\end{abstract}

\begin{graphicalabstract}
\end{graphicalabstract}

\begin{highlights}
\item Model calibration is more important than accuracy for sports betting
\item Sports bettors can increase their wealth by a third over a single season
\item Kelly betting only works with a well-calibrated model
\end{highlights}


\end{frontmatter}


\section{Introduction}
\label{}
Sports betting in the US is conservatively estimated to be a \$150 billion industry \cite{legalsportsbetting}. As a result of the worldwide interest in US sports, an abundance of data is publicly available. Much research has been done on the use of machine learning (ML) for sports outcome prediction. The success of this research has turned this data into an invaluable commodity for sports bettors around the world. If a bettor can leverage data to accurately estimate the true probability of a sporting outcome, they can identify a bookmaker's mispricing of this outcome - and whether or not there is an opportunity to make a profit. The development of a proficient predictive model could therefore prove extremely lucrative.

The National Basketball Association (NBA) in North America is the world's premier basketball league. This paper focuses on the development of a data-driven betting system in the case of the NBA. ML for sports outcome prediction has been widely studied, however very little of this research extends to sports betting. In the bulk of this work, ML models are evaluated on accuracy achieved. Accuracy is a suitable model evaluation metric for the sports outcome prediction problem, as the goal is to correctly predict the winning team \cite{bunker2019machine}. However this is not necessarily true for the sports betting problem, where the goal is to estimate the true probability of the
sporting outcome to identify and profit from any mispricing of the odds offered on this outcome. As \textit{calibration} is used to estimate how close a model's predicted probabilities are to the true probabilities, we hypothesise that calibration is a more appropriate metric than accuracy for the sports betting problem, and that basing model selection on calibration, rather than accuracy, leads to greater profit generation.

The simplest way to test this hypothesis is to consider the most straightforward form of a wager – the moneyline bet. To win a moneyline bet, the bettor must predict the winner of the game \cite{hubavcek2019exploiting}. If the bettor correctly predicts the winner, they win back the wager (also called the stake) along with the profit, otherwise, they lose the wager to the bookmaker \cite{hubavcek2019exploiting}. The profit is a predetermined quantity that corresponds to the odds. Decimal odds display the total return (stake plus profit) that a wager of a single unit could yield \cite{cortis2015expected,cortis2016betting}. Taking the inverse of the odds results in the \textit{implied probability} of the outcome occurring \cite{cortis2015expected,cortis2016betting}. Let us assume that $\pi_{i}$ represents the bookmaker's implied probability of outcome $i$. 
Then the odds are given as $1/\pi_{i}$.

For fair odds, the implied probability represents the bookmaker’s estimate of the probability of that outcome occurring \cite{hubavcek2019exploiting}. In practice, the odds are never fair, as the bookmaker wants to ensure they make a profit \cite{wheatcroft2020profiting}. The odds deviate from the 'fair price' by the bookmaker's \textit{margin}. This is the absolute difference between 1 and the sum of the implied probabilities, and can be viewed as a commission charged to bettors by the bookmaker \cite{hubavcek2019exploiting}. Unsurprisingly, each possible outcome of a coin flip is equally likely, so the fair odds for each outcome should be 2.0. A bookmaker might offer odds of 1.90 for each so the sum of implied probabilities would be $(1/1.90) + (1/1.90) \approx{1.0526}$, which suggests that the bookmaker has factored in a margin of approximately 5\%. When the true probability of an outcome occurring is greater than the probability implied by the bookmaker's odds, the expected value of the bet is positive. We define such a bet as a value bet \cite{edwards1955prediction}. Value bets arise when the bookmaker misprices the odds. To spot such opportunities, bettors can compare their model's predicted probability to the bookmaker's odds. We follow this approach and implement and evaluate betting systems which aim to identify value bets and capitalise on them. We measure the success of a system by the return on investment (ROI) achieved over the course of an NBA season.

The predictive model is perhaps the most important part of a sports betting system, and there are many algorithms available to bettors \cite{zdravevski2010system,cao2012sports,zimmermann2013predicting,alonso2022machine,cheng2016predicting,tran2016predicting,pai2017analyzing}. Prominent examples used in research include support vector machines (SVM), logistic regression (LR), Naive-Bayes, K-nearest neighbours (kNN), decision trees and neural networks \cite{hamadani2006predicting,miljkovic2010use,loeffelholz2009predicting}. Although many predictive algorithms have already been explored, there has still been a distinct lack of exploration of model evaluation metrics. In the vast majority of this research, models have been evaluated based on accuracy alone. As accuracy may not be the most appropriate metric for the sports betting problem, the lack of alternative metrics to evaluate model performance is a key gap in the literature. Our hypothesis, that calibration is a more suitable metric in this setting, addresses this gap in the literature. We therefore design competing betting systems, basing model selection on accuracy in one system, and calibration in the other. Comparing the returns achieved by each betting system, we can determine whether using accuracy or calibration as the basis for model selection leads to greater profit generation. One of the interesting findings of our work is that, on average, and in the best case scenario, selecting the predictive model based on calibration, rather than accuracy, leads to greater profits.

The remainder of this paper is organised as follows. Related works are discussed in section 2. section 3 involves a statement of the central hypothesis of the paper and explores the reasons behind the authors' arrival at this hypothesis. section 4 lays out the design of an experiment to test this hypothesis. section 5 covers the novel feature engineering carried out ahead of the predictive modelling, which is covered in section 6. section 7 discusses the betting experiments and the algorithm used to conduct the betting simulations in detail. In section 8, results of the experiments are examined. Finally, section 9 discusses the implications of our findings and concludes the paper.

\section{Related Work}

While analysing data from the National Football League (NFL) to understand bettor and bookmaker strategies, Levitt and colleagues identified various approaches used by bookmakers to generate profits \cite{levitt2004gambling}. The first approach relies on the bookmaker's ability to anticipate the price which equalises the quantity of money wagered on each side of the bet \cite{levitt2004gambling}. If done successfully, the losers compensate the winners while the bookmaker collects the margin. If the bookmaker doesn't get this right they risk having to reach into their own reserves to compensate the winners. A second approach involves the bookmaker being able to systematically outperform bettors in game outcome prediction. This allows the bookmaker to set the 'correct' price. While the money wagered is not equalised on any given game, this approach sees the bookmaker profit off the margin on average over the course of the season \cite{levitt2004gambling}. To increase profits, the bookmaker may combine the previous two approaches and set the 'wrong' price on purpose to exploit bettor preferences. However, if the odds deviate too much from the true price, shrewd bettors aware of the 'correct' price can capitalise on this, and make a profit  \cite{levitt2004gambling}. Levitt found that bookmakers generally focus on outperforming the average bettor in outcome forecasting \cite{levitt2004gambling}. This leaves bettors with an opportunity to generate positive returns if they can identify when the bookmaker's price is wrong.

In order to identify value bets, bettors must first possess a reliable predictive model. As the aim of sports outcome prediction is to predict the outcome of a sporting event given a finite set of possibilities, it is usually addressed as a classification problem \cite{horvat2020use}. To quantify the performance of classifiers in such settings, accuracy is generally used as the model evaluation metric, where accuracy refers to the proportion of correctly classified data \cite{horvat2020use,yang2020hyperparameter}.  In their work, Bunker and Thabtah deemed this appropriate, noting 'classification accuracy is a reasonable measure of evaluation' for the sports outcome prediction problem \cite{bunker2019machine}. Many different classifiers have been used to address this problem with neural networks among the most widely used \cite{horvat2020use}. Other classifiers commonly used for the sports outcome prediction problem include SVM and kNN, both of which are widespread in baseball forecasting \cite{zhang2000neural,horvat2020use}. When it comes to designing a robust sports prediction model, other key considerations one must take into account include the features to use, as well as methods of evaluating the model's performance. A common approach for model evaluation is chronological data segmentation, i.e. using a training set made up of seasons prior to those in the evaluation set  \cite{horvat2020use,horvat2019importance}. For sports prediction, it is critically important to preserve the chronological order of the training data. This is done to ensure that upcoming matches are predicted based on data from past matches only. As cross-validation usually involves shuffling the order of the instances, it is not recommended in the sports prediction setting \cite{bunker2019machine,horvat2020use}. In general, feature selection and feature extraction are also employed to reduce the dimensionality and complexity of the classification problem \cite{horvat2020use}.

Much of this work has focused on basketball, likely due to the abundance of publicly available data \cite{BasketballReference, nba, databasketball,basketballgeek}. Researchers have made efforts to identify the most important features and best-performing classifiers. Notably, Ivankovic found the most important features to be, in order of diminishing importance, defensive rebounds, two-point and three-point shots, steals, turnovers, offensive rebounds, free-throw shots, blocks and assists \cite{ivankovic2010analysis}. In terms of identifying the best performing classifiers, Cao and colleagues trained several models on NBA data from the 2005/2006-2009/2010 seasons, and evaluated these models on the 2010/2011 season \cite{cao2012sports}. In this study, LR was found to be the best-performing model with an accuracy of 69.97\%, outperforming multi-layer perceptron (MLP) and SVM models (which achieved accuracies of 68\% and 67.7\%, respectively) \cite{cao2012sports}. In contrast, Torres found an MLP model to be superior to LR in the same setting, with accuracies of 68.44\% and 67.98\%, respectively \cite{torres2013prediction}. In a separate study, Lin et al discovered that a team's win/loss record is a crucial predictor of victory \cite{lin2014predicting}. Leveraging the impressive predictive capability of neural networks, Hubá\v{c}ek and colleagues constructed a neural network that used a convolutional layer to summarise player-level features into team-level features. By considering only high-confidence predictions, this classifier achieved an accuracy of 84.35\%, compared to an accuracy of 80\% achieved by a neural network that only used team-level features. Despite the success researchers have achieved, practical implementation of such forecasting systems poses several limitations, as pointed out by Ganguly and Frank, including $(i)$ lack of context, $(ii)$ no measure of uncertainty of the prediction and $(iii)$ lack of benchmark datasets to compare results \cite{ganguly2018problem}. Another common limitation is that many predictive models do not take in-game events into account. Events such as the early injury of a star player can have a significant influence on the outcome of a game \cite{horvat2020use}. Despite this limitation, we focus on pre-game betting based on the closing odds. Naturally, considering in-game events would require the ability to process streaming data, whereas generating predictions for pre-game betting can be done using batch processing, and resources for batch processing are much more readily available to the average bettor \cite{pfandzelter2019iot}.

For a gambler seeking to maximise profits over a series of successive bets (with the opportunity to reinvest the winnings), the size of each bet can be determined using the \textit{Kelly criterion} \cite{kelly2011new}. The criterion identifies the optimal bet size by maximising the expected value of the logarithmic growth of wealth \cite{7447000}. While its origins lie in the analysis of long-distance telephone signal noise, the equation has found widespread application across many domains \cite{dotan2020beating,rotando1992kelly,thorp2008kelly,thorp1975portfolio,barnett2010applying}. Sports bettors can use this criterion to calculate optimal bet size as a proportion of overall bankroll \cite{jacot2023kelly}. They simply require the bookmaker's odds, and the probability of victory according to their predictive model. Mathematically, the Kelly criterion can be defined as shown in (1): 

\begin{equation}
    k = \frac{pb-q}{b} 
\end{equation}

where:
\begin{itemize}
    \item $k$ is the proportion of the bettor's bankroll to wager on the given outcome
    \item $p$ is the probability of the given outcome occurring
    \item $b$ represents the potential winnings on a wager of 1 unit i.e. $odds - 1$
    \item $q$ is the probability of the given outcome not occurring, i.e. $q=1-p$ \cite{dotan2020beating} 
\end{itemize}

Despite the criterion's reputation as the optimal strategy for resource allocation on a set of gambles repeated over time, relying on it to decide bet size has certain limitations \cite{7447000}. For instance, the criterion often suggests wagering a very large proportion of the overall bankroll on a single game, which is a recipe for disaster in a realm as unpredictable as the world of sport. The Kelly strategy in this form therefore leads to almost sure ruin \cite{7447000}. In contrast, the \textit{fractional Kelly} is a less risky variation of the strategy. This variation uses the same formula, but here $k$ represents the proportion of a \textit{fraction} of the overall bankroll. Thus, implementing the quarter-Kelly would mean for $k=0.4$, instead of wagering $40\%$ of their bankroll, the bettor wagers $10\%$. Recently, Dotan illustrated the utility of the fractional kelly in NBA betting markets. Notably, in a betting simulation spanning a single season, the '$5^{th}$ Kelly' strategy earned an ROI of over 98\%, while its full-Kelly counterpart crashed to zero \cite{dotan2020beating}.

Excluding accuracy, the lack of metrics used to evaluate the performance of sports prediction models is a noticeable gap in the literature \cite{horvat2020use}. While it may be suitable for the sports outcome prediction problem, accuracy alone is not a sufficient model evaluation metric for the sports betting problem. To compensate for this, Hubá\v{c}ek and colleagues designed a loss function to penalise correlation with the bookmaker's odds \cite{hubavcek2019exploiting}. One of the key findings of their work was that training models under this loss resulted in greater profits than optimising for accuracy \cite{hubavcek2019exploiting}. Further, it has been observed that a highly accurate predictive model is useless as long as it coincides with the bookmaker’s model \cite{hubavcek2019exploiting}.

Combining these findings with the aforementioned gap in the literature leads to the central hypothesis of this paper, which is discussed in the next section.

\section{Central Hypothesis}

The purpose of a sports betting model is to predict the probability of victory for each team in a given game, so that these probabilities can be compared to the bookmaker’s odds to determine if a value bet is on offer. The model may also play a role in deciding how much to bet, e.g. if the bettor makes use of the Kelly criterion to decide bet size. Therefore, for a betting system to be successful, it is critically important that the probability of victory generated by the model is close to the true probability of victory. 
This notion, that the probability a classifier assigns to an event should reflect the true frequency of that event, relates to the concept of \textit{calibration} \cite{kumar2019verified}. In contrast to calibration, the fundamental problem with accuracy in this setting is that it does not take into account the distance between the predicted probability of victory and the true probability of victory. Accuracy simply measures the proportion of correctly classified data. Instead of accuracy, we desire a metric that provides an indication of the distance between the predicted probabilities and the true probabilities.

To discuss calibration formally, let us consider the problem of multiclass classification. Suppose we have input $X \in{\mathbb{X}}$  and label $Y \in{\mathbb{Y}} = \{1, . . . ,K\}$ which are random variables with
ground truth joint distribution $\pi{(X, Y )} =
\pi{(Y |X)}\pi{(X)}$. Then our classifier is of the form $f(X) =
(\hat{Y} , \hat{P})$, where $\hat{Y}$ is the predicted label and $\hat{P}$ is the probability associated
with the prediction. 
The probabilistic classifier is said to be well-calibrated if, among test instances assigned a predicted probability
vector $\hat{P}$, the class distribution is (approximately) distributed as $\hat{P}$ \cite{kull2019beyond}. Perfect calibration occurs if

\begin{equation}
\mathbb{P}(\hat{Y}=Y | \hat{P}=p)=p, \forall{p \in{[0,1]}}    
\end{equation}

with reference to the probability over the joint distribution \cite{guo2017calibration}.

Many methods have been proposed to measure calibration. We use the classwise expected calibration error (classwise-ECE), as this variation overcomes some of the limitations of the original ECE \cite{kull2019beyond}. To calculate the classwise-ECE, the interval $[0,1]$ is split into $M$ bins of equal length so that the $m^{th}$ bin is the interval $[\frac{m-1}{M},\frac{m}{M})$. For a given class $k$, each prediction in the set is grouped into the bin its probability lies within, i.e. we associate $\hat{P}_{k}(x)$ with the $j^{th}$ bin if $\hat{P}_{k}(x) \in{[\frac{j-1}{M},\frac{j}{M})}$. Let $B_{j,k}$ represent the $j^{th}$ bin for predicted probabilities relating to class $k$. The classwise-ECE is defined as shown in (3):  

\begin{equation}
classwise-ECE = \frac{1}{k}\sum_{i=1}^{k}\sum_{j=1}^{m}\frac{|B_{j,i}|}{n}|y_i(B_{j,i})-\bar{p}_{i}(B_{j,i})|     
\end{equation}

where $k$ is the number of classes in the problem, $m$ is the number of bins used, $n$ is the size of the dataset, $|B_{j,i}|$ is the size of the bin (number of class $i$ predictions associated with the $j_{th}$ bin), $y_i(B_{j,i})$ is the actual rate of occurrence of class $i$ in $B_{j,i}$ and $\bar{p}_{i}(B_{j,i})$ denotes the average probability of class $i$ predictions in $B_{j,i}$ \cite{kull2019beyond}. This error is bounded between 0 and 1, and can be thought of as the percentage by which the model's predicted probability deviates from the true probability, on average. We note that for binary classification, classwise calibration is equivalent to class-specific calibration focusing on the positive class, since the predicted probability of the negative class is
determined by the predicted probability of the positive class \cite{posocco2021estimating}.
We can equivalently state that class $i$'s contribution to the classwise-ECE is equal for both classes in the binary case (due to symmetry), and so we can simply calculate the error for just the positive class, as this is equivalent to the average error of the two classes.

One of the limitations of the original expected calibration error is that it focuses only on the probability of the most likely class, and for each prediction, ignores calibration with respect to the $K-1$ other classes \cite{nixon2019measuring}. While the classwise-ECE overcomes this, a concerning limitation still exists - one can obtain almost perfectly calibrated probabilities by predicting the overall class distribution for all instances \cite{kull2019beyond}. To avoid this scenario, we impose a constraint on model predictions - we require at least 80\% of the bins (that the predicted probabilities are grouped into) to be non-empty. This ensures that the distribution of predictions is not too highly concentrated around the mean.

Having introduced the concept of calibration, we arrive at the central hypothesis of this paper - for the sports betting problem, it is more important for a classifier to be well-calibrated than highly accurate. Therefore, basing model selection on calibration, rather than accuracy, should allow for greater profit generation. We design an experiment to test this hypothesis, as discussed in the next section.

\section{Experiment Design}

A data-driven sports betting system that selects its predictive model on the basis of calibration should generate greater profits than an identical system that selects its model based on accuracy. This is the idea at the core of this paper. To test this notion, we design an experiment in which two sports betting systems compete, one basing model selection on accuracy, the other on calibration. In this experiment, the model selection paradigm extends beyond selection of the optimal learning algorithm, to include feature selection and selection of optimal hyperparameter values, due to the significant contributions these processes make to model performance \cite{binder2020multi}. We begin with a set of candidate learning algorithms (logistic regression, random forest, support vector machines, multi-layer perceptron). Next, we construct a predictive modelling pipeline (see Fig. 1) that carries out feature selection and hyperparameter-optimisation, prior to selecting the best-performing model. This pipeline consists of two branches, one that evaluates performance based on accuracy, the other based on calibration. Along the accuracy branch we aim to maximise accuracy, and along the calibration branch we aim to minimise the classwise-ECE (using 20 bins). Finally, the models are evaluated on a test set to select the best-performing model under each metric. The final output of the pipeline consists of a model from each branch - the most accurate model from the accuracy branch, and the most well-calibrated model from the calibration branch.

\newcommand*{\MyIndent}{\hspace*{0.5cm}}
 
\begin{figure*}[ht]
    %\hspace*{-0.75cm}
    \centering
    \includegraphics[scale=0.65]{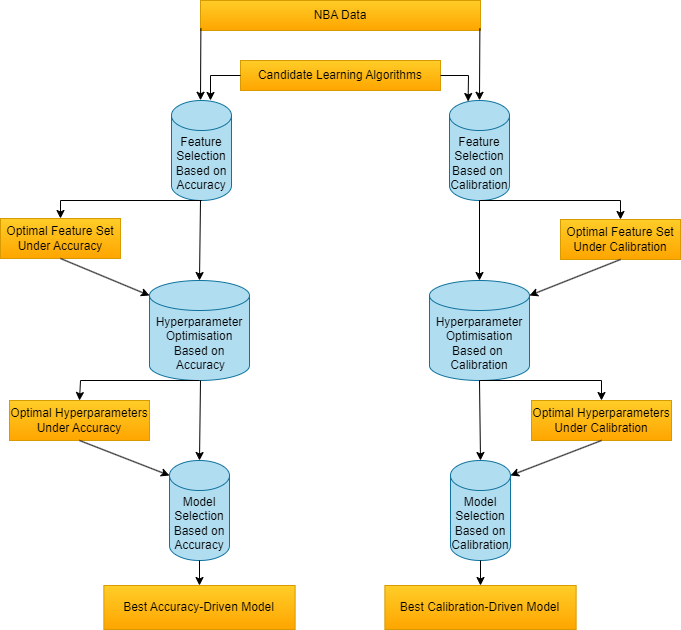}
    \caption{Predictive Modelling Pipeline: Two branches of the pipeline are presented. The first branch employs model selection (including feature selection and hyperparameter optimisation) using accuracy as the evaluation criterion, prior to selecting the most accurate model. The second branch employs model selection based on calibration, selecting the most well-calibrated model. These branches take NBA data and a set of candidate learning algorithms as inputs. The output of each branch is the optimal predictive model under the given metric. Here the blue cylinders represent ML processes and the yellow rectangles represent inputs and outputs of these processes.}
\end{figure*}

\newpage

The two final models are used to generate predictions for each game in an NBA season. We then implement separate betting systems for each set of predictions. Typically, a betting system has two components: a strategy and a rule. 
\begin{itemize}
    \item The betting strategy decides whether or not to place a bet 
    \item The betting rule decides the size of the bet \cite{dotan2020beating}
\end{itemize}

For each game, these decisions are made by comparing the model's predicted probabilities to the bookmaker's odds. Ultimately, the betting systems are evaluated by their ROI, where the ROI is the percentage change in the bettor's initial bankroll by the end of the season.

\begin{figure*}[t]
    \centering
    \includegraphics[scale=0.65]{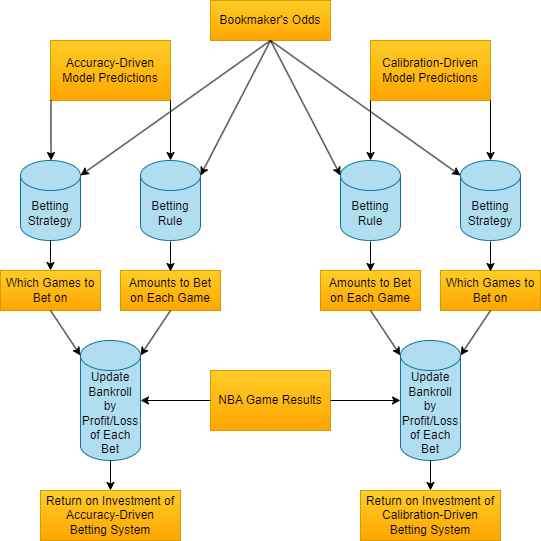}    
    \caption{Betting Simulation Pipeline: Two branches of the pipeline are presented. The first branch represents the simulation for a bettor using a predictive model selected on the basis of accuracy, while the second branch represents the simulation for a bettor using a predictive model selected on the basis of calibration. These branches take as inputs a set of predictions, the bookmaker's odds and the results of each game in the given NBA season. For a given combination of strategy and rule, the output of each branch is the return on investment achieved by the bettor. The meaning of each shape and colour is given in the caption of Fig. 1.}
\end{figure*}

This experiment is designed to answer the following question:
in a data-driven sports betting system, does model selection based on calibration, rather than accuracy, allow for greater profit generation?
To answer this, we examine the ROI achieved by each betting system, to determine whether or not basing model selection on calibration leads to greater profits.

\smallskip 

A crucial, preliminary step of the experiment is feature engineering.

\newpage
\section{Feature Engineering}

NBA games cannot end in a draw - in the case of a tie, successive overtime periods are played until there is a winner. This makes predicting the outcome of NBA games a binary classification problem where the aim is to predict the winning team. Many researchers have successfully applied supervised learning to this problem, largely constructing their features using box score statistics. A comprehensive list of basic and advanced box score statistics is provided in the supplementary document (see Tables 1 and 2).

A common approach to construction of the feature set is to use box score statistics averaged over the season to date as predictor variables for each game (e.g. average blocks per game) \cite{hubavcek2019exploiting}. This approach considers how well a team performs on average in a particular aspect of the sport. However, considering absolute figures like this can leave the data prone to shift, as the characteristics of the league may change over time \cite{dutta2017identifying}. Therefore, we average the differences in team-total box score statistics versus previous opponents over the season to date. This approach considers the amount by which a team \textit{outperforms} their opponent on average in each particular aspect of the sport, as these relative figures are less prone to shift \cite{dutta2017identifying}. We then perform feature extraction by taking the difference of the home and away teams' 'average out-performance values' for each box score statistic, to reduce the dimensionality of the data. To demonstrate the construction of such features, we provide a hypothetical calculation using synthetic data in the supplementary document (see Tables 4 and 5).

Interestingly, Lin and colleagues found that box score statistics alone may not provide enough information for accurate prediction, and concluded that using a team's win/loss record can improve accuracy \cite{lin2014predicting}. Therefore, we include another feature in our model - each team's regular season winning percentage from the previous year. For each game, we take the difference between the home and away team's winning percentages from the previous season, and denote this feature 'Previous Season Winning Percentage'. To ensure each prediction is sufficiently well-informed, we exclude the first 10 games of each team in each season from the training instances, and use them only in the calculation of later games' features \cite{hubavcek2019exploiting}.

Our final feature engineering step is feature standardisation. For each feature, this involves subtracting the mean and dividing by the standard deviation so that it is distributed with a mean of 0 and standard deviation of 1 \cite{labayen2020online}. This is done to ensure all features are on the same scale, as models that are smooth functions of the input are affected by the scale of the input, and we do not want the range of a feature's values to dictate the influence it has on the model \cite{zheng2018feature}. Generally, both the training and test set features are scaled using the distribution of the training set. This approach assumes these distributions are approximately equal. If this is not the case (a phenomenon known as covariate shift), it may yield inaccurate results \cite{sugiyama2007covariate}. To detect covariate shift, we compare the distribution of each feature in a validation set to its distribution in an initial training set which is composed of the seasons prior to the validation set. If these are found to be different, the feature is dropped. To test if two samples are drawn from the same (unknown) distribution, the two-sample Kolmogorov-Smirnov test is used \cite{pratt2012concepts}. We carry out these tests at the 1\% level of significance, as strong evidence of covariate shift is necessary for a feature to be dropped. All remaining features are then standardised - using their own distribution for training sets, and using the distribution across all prior seasons for validation, test, or betting simulation sets. With feature set constructed, we have finalised the NBA data required by our predictive modelling pipeline. The pipeline is discussed in detail in the next section.

\section{Predictive Modelling}

The final input component of the predictive modelling pipeline is the set of candidate predictive models. We explore these probabilistic classifiers in detail below.

\subsection{Candidate predictive Models}
The first candidate predictive model is LR. Suppose we have a dataset with $n$ samples $D = \{(X_1,y_1),...,(X_n,y_n)\}$, where each sample has $d$ features i.e. $X_i = (X_{i1},...,X_{id})$ and a corresponding response $y_i$. Given input $X_i$, LR gives the probability of the sample belonging to class $C$ by (4).

\begin{equation}
P(Y_i=C|X_i) = f(X_i\beta)=  \frac{\exp{(X_{i}\beta)}}{1+\exp{(X_{i}\beta)}}    
\end{equation}

where $\beta = (\beta_{0},\beta_{1},...,\beta_{d})$ are the model parameters to be estimated (with intercept $\beta_0$) \cite{liang2013sparse}.

The log-likelihood is shown in (5).

\begin{equation}
L(\beta | D) = - \sum_{i=1}^{n}\{y_i log[f(X_i\beta)] + (1-Y_i)log[1-f(X_i\beta)]\}     
\end{equation}

LR learns $\beta$ that minimises this.

The next candidate is the random forest (RF) algorithm, an ensemble learning method which
makes use of bagging to combine multiple decision trees (DT) \cite{injadat2018bayesian}. RF involves the use of bootstrapping to randomly generate various sub-samples of the dataset, before fitting a decision tree to each \cite{sklearn}. To classify a sample, each tree evaluates it individually, and the class which receives the most votes is selected as the final classification result \cite{salo2019clustering}.

This is followed by the SVM algorithm. SVMs are supervised learning
models which can be used for classification and regression \cite{smola2004tutorial}. They work by partitioning data points using hyperplanes as decision boundaries, and often map data into higher-dimensional space to make the points linearly separable \cite{yang2018image}.

For a dataset of size $n$, the objective function is given by (6) \cite{zhang2003modified}.
 
\begin{equation}
L = arg min_{w} \Bigg\{ \frac{1}{n} \sum_{i=1}^{n} max\{0,1-y_{i}f(x_{i})\} + Cw^Tw
\Bigg\}    
\end{equation}

where $w$ is a normalisation vector and $C$ is a regularisation hyperparameter.
$f(x)$ is a function which measures the similarity between two points, known as the kernel.
This function can come in the form of a radial basis function (RBF), linear kernel, polynomial
kernel, or sigmoid kernel.

The final candidate is the MLP. This is a layered, feed-forward neural network where each layer is composed of nodes with each node connected to every other node in the subsequent layer \cite{delashmit2005recent}.

Each of these models takes NBA data as input, and returns the predicted probability of each team winning, for a given game. To ensure reliable predicted probabilities, our features must be well-selected. We discuss the feature selection process below.

\subsection{Feature Selection}

Feature selection (FS) refers to the detection of relevant features and removal of irrelevant, redundant, or noisy data \cite{kumar2014feature}. Various studies have shown the ability of FS to minimise the dimensionality of the problem, maximise the accuracy of classification and prevent overfitting \cite{wah2018feature,li2017feature,kira1992practical}. Of the many FS methods available, we make use of a filter method and a wrapper method in our pipeline. $(i)$ Filter methods are those used to evaluate the relevance of features that are independent of the learning algorithm, e.g. ranking of features based on correlation with the response variable \cite{kumar2014feature}. $(ii)$ Wrapper methods are those which evaluate candidate feature subsets (under the given evaluation criterion) using a learning algorithm, and select the best-performing subset \cite{kumar2014feature}.

$(i)$ Two features are considered highly correlated if their spearman correlation coefficient is greater than 0.7 \cite{xiao2016using}. 
 For a given group of highly correlated features, we consider all except the feature which is most correlated with the target to be redundant, and so remove them. This initial step is common to both branches of the predictive modelling pipeline (see Fig. 1), as filter methods are generally considered a pre-processing step
\cite{kotsiantis2006data}. The features remaining after application of the filter method are denoted as subset A.

$(ii)$ Next, a wrapper method is applied. Sequential forward selection (SFS) is used with LR as the learning algorithm and completion of search as the stopping criterion. This is implemented as a separate process along each branch, and for each, the input to the process is feature subset A. For the calibration branch, the feature subset under which the LR model achieves the lowest classwise-ECE (fitted to an initial training set and evaluated on a validation set) is considered the optimal feature subset for calibration and is denoted as subset B. This subset of features is used for all further modelling along the calibration branch. For the accuracy branch, the feature subset under which the highest accuracy is achieved is considered optimal and is denoted as subset C. Subset C is used for all further modelling along the accuracy branch.

Due to the limitation of the classwise-ECE discussed in section 3, we place a constraint on model predictions. After grouping predictions into the corresponding bins, we look at the distribution of the weights of each bin, where the weight of a bin refers to the proportion of predictions associated with it. We want to ensure the predictions (for a given class) are not concentrated in a small number of bins. To prevent such scenarios, we impose the following constraint on model predictions: if the distribution of bin weights is such that less than 80\% of the bins are non-empty, we set the classwise-ECE to 1 (the maximum possible error).  This is to ensure that we do not allow models to achieve a low classwise-ECE by generating predictions that are approximately equal to the overall class distribution for all instances. Without imposing this constraint, we could mistakenly identify a sub-optimal feature set as the optimal feature set. This constraint is also applied to model predictions during the hyperparameter optimisation and model selection processes. Naturally, this constraint does not apply to the accuracy branch.

\subsection{Hyperparameter Optimisation}

Model performance is significantly influenced by the choice of hyperparameters, and automating the process of hyperparameter tuning has become the focus of much research in recent years. Automated hyperparameter optimisation (HPO) has several important benefits, including reduction of human effort required for applying ML, improvement in performance of ML models, and improvement in reproducibility and fairness of research \cite{hutter2019automated}. Many optimisation problems are non-convex or non-differentiable, in which case traditional optimisation techniques may result in a local rather than global optimum \cite{luo2016review}. Popular among the non-traditional optimisation techniques that have been used for HPO problems is an iterative algorithm known as bayesian optimisation (BO) \cite{snoek2012practical}. BO is considered more efficient than traditional HPO techniques like random search or grid search, because the algorithm decides which points in the hyperparameter search space to evaluate based on previously-obtained
results, rather than letting the user specify the points \cite{hazan2017hyperparameter}. We implement a form of BO known as BO-TPE, regarded as one of the most suitable HPO techniques for LR, RF, SVM, and MLP classifiers \cite{yang2020hyperparameter}.

For a specified predictive model, along with hyperparameters to be optimised, the search space for each hyperparameter, training and validation data to train and evaluate the model, and an objective function to be minimised, the algorithm identifies the optimal set of hyperparameter values (and returns the corresponding score on the validation data). Due to the element of inherent randomness in this process, we run the algorithm several times over different random seeds before selecting the optimal set of hyperparameters. The full process is described in the supplementary document. HPO is implemented separately along each branch of the predictive modelling pipeline (see Fig. 1). For the calibration branch, the objective function is the classwise-ECE, while the negative of accuracy is used for the accuracy branch. The hyperparameter search space for each model is given in the supplementary document (see Table 7).

\subsection{Model Selection}

The final stage of the predictive modelling pipeline is model selection. We fit each model to an extended training set (consisting of the initial training data combined with the validation data) under the optimal feature set and hyperparameter values for the given branch, and generate predictions for the test set. We then evaluate these predictions under the given metric. Along the calibration branch, the candidate predictive model that achieves the lowest classwise-ECE on the test set is deemed to be the best calibration-driven model. Along the accuracy branch, the model which achieves the highest accuracy on the test set is selected as the best accuracy-driven model. These two models are the final output of the predictive modelling pipeline, and are used to generate predictions for each game in an NBA season. These predictions (combined with the bookmaker's odds) are the input of our betting experiments, as detailed in the next section.

\section{Betting Experiments}

We fit each model selected by the pipeline to a final training set, and generate predictions for a betting simulation set. The final training set comprises the extended training and test sets, and the betting simulation set consists of a single NBA season (details of each data set are discussed in section 8). These two sets of predictions are used to implement competing betting systems, over the given season.

We carry out the betting simulation for a given system as follows. Beginning with an initial bankroll of \$10,000, we iterate through each game in the betting simulation set in chronological order, and for each:

\begin{enumerate}
    \item For each team, we compare the model's predicted probability of victory to the probability implied by the bookmaker's odds, to decide whether or not to place a bet. This decision is determined by the betting strategy 
    \item If the decision is to bet, we determine the stake by the betting rule
    \item We subtract the stake from the bankroll
    \item If the bet is successful, we add the stake and the winnings to the bankroll
\end{enumerate}

Both strategy and rule are crucial elements of the betting system. While many possibilities could be explored, this is outside of the scope of this paper. Our goal is to determine whether basing model selection on calibration, rather than accuracy, leads to greater returns, and we do not want differences in ROI achieved by each system to be attributed to other factors such as strategy used, etc. Therefore, we use only the simplest possible strategy and rules.

\subsection{Strategy}

We implement the simplest possible strategy: bet on all value bets identified by the model (i.e. bet if a team's predicted probability of victory is greater than the probability implied by the bookmaker's odds).

\subsection{Rules}

We implement two simple rules. The first is fixed betting - each time we decide to bet, we set the stake to \$100. The second involves the Kelly criterion - each time we decide to bet, we determine the stake using the eighth-Kelly \cite{7447000}. As previously discussed, the full-Kelly is considered too aggressive, and leads to almost sure ruin \cite{dotan2020beating,7447000}. The eighth-Kelly is a conservative alternative, that recommends betting an eighth of the optimal bet size. This represents a more realistic scenario than fixed betting, as both the odds offered and the bettor's perceived probability of the outcome generally influence the choice of stake \cite{Matej_2021,jacot2023kelly}.

The algorithm used to conduct each betting simulation is shown in Table 1.

\begin{table}[t]
    \centering
    \caption{Betting simulation algorithm}
    \begin{tabular}{p{8.5cm}}
     \hline
    Bankroll = $\$10,000$\\
    \\
    For each game in the dataset:\\
    \MyIndent $P_h$ = Predicted probability of victory for home team\\
    \MyIndent $O_h$ = Bookmaker's odds for home team victory \\
    
    \MyIndent $P_a$ = Predicted probability of victory for away team\\
    \MyIndent $O_a$ = Bookmaker's odds for away team victory \\
    \\
    \MyIndent If $P_h > 1/O_h$:\\
    \MyIndent \MyIndent $K$ = ($P_h \times$ Bankroll $- (1-P_h)$)/Bankroll  \\
    \MyIndent \MyIndent Stake = $  \begin{cases}
    1/8 \times K \times Bankroll,& \text{if Rule = Eighth-Kelly} \\
    \$100,              & \text{if Rule = Fixed Betting}
\end{cases}  $ \\
    \MyIndent \MyIndent Bankroll = Bankroll - Stake \\
    \MyIndent \MyIndent If home team wins:\\
    \MyIndent \MyIndent \MyIndent Winnings = (Stake $\times O_h$) - Stake \\
    \MyIndent \MyIndent \MyIndent Bankroll = Bankroll + Stake + Winnings\\ 
    \\
    \MyIndent Else If $P_a > 1/O_a$:\\
    \MyIndent \MyIndent $K$ = ($P_a \times  $Bankroll $- (1-P_a)$)/Bankroll  \\
    \MyIndent \MyIndent Stake = $  \begin{cases}
    1/8 \times K \times Bankroll,& \text{if Rule = Eighth-Kelly} \\
    \$100,              & \text{if Rule = Fixed Betting}
\end{cases}  $ \\
    \MyIndent \MyIndent Bankroll = Bankroll - Stake \\
    \MyIndent \MyIndent If away team wins:\\
    \MyIndent \MyIndent \MyIndent Winnings = (Stake $\times O_a$) - Stake \\
    \MyIndent \MyIndent \MyIndent Bankroll = Bankroll + Stake + Winnings\\     
    \\
    \MyIndent Until end of dataset is reached\\    
     \hline
    \end{tabular}    
\end{table}

As discussed, each betting system consists of a strategy and a rule, in addition to a predictive model selected on the basis of either calibration or accuracy. Following the algorithm described in Table 1, we simulate each betting system over a single NBA season. Measuring the ROI achieved by each system, we compare the profitability of calibration-driven systems to their accuracy-driven counterparts, to test our hypothesis. The results of this experiment are discussed in the next section.

\section{Results}

To undertake this research, we obtained NBA data from the 2014/2015-2018/2019 seasons from basketball-reference.com \cite{BasketballReference} We used the 2014/2015-2015/2016 seasons as an initial training set. The models were fitted to this data during the FS and HPO processes. The 2016/2017 season was used as a validation set to evaluate model performance during these processes. After the FS and HPO processes were completed, the 2014/2015-2016/2017 seasons were used as an extended training set. The predictive models were fitted to this data ahead of model selection, during which they were evaluated on a test set consisting of the 2017/2018 season. The best-performing models were then fitted to a final training set spanning the 2014/2015-2017/2018 seasons and used to generate predictions for the 2018/2019 season - which comprised our betting simulation set. A key requirement for the betting simulation was obtaining authentic odds published by a bookmaker. To fulfill this requirement, we obtained the publicly available closing moneyline odds for the 2018/2019 NBA season, published by Las Vegas sportsbook Westgate \cite{sportsbookreviewsonline}.

Certain features were dropped from the dataset prior to FS, as a result of showing signs of covariate shift, being a linear combination of other features, or presenting only null values. A list of these features is provided in the supplementary document (see Table 3). Many more features were considered redundant by each of the FS methods employed. The subsets generated by each of the FS methods, along with the features dropped and features selected by each method are provided in (see Table 6). Subset B, the optimal feature subset for calibration-driven predictive modelling, consisted of the following features: 3P, BLK, FT\%, ORB, AST\%, Previous Season Winning Percentage. Subset C, the optimal feature subset for accuracy-driven predictive modelling, consisted of the following features: STL\%, eFG\%, DRB, AST, DRtg, ORtg, 3P, Previous Season Winning Percentage.

Next, HPO was implemented along each branch using the BO-TPE algorithm. The optimal hyperparameter values identified for each model are provided in the supplementary document (see Tables 8 and 9). These values were used for all further modelling steps.

Using the optimal feature set and hyperparameter values identified for each branch, the models were evaluated on a test set. Their scores are provided in Tables 2 and 3.

\begin{table}[h]
    \centering
    \caption{Classwise-ECE achieved on the test set by each model along the calibration branch}
    \begin{tabular}{ |p{4.5cm}|p{2.8cm}| }
     \hline
     \textbf{Model} & \textbf{Classwise-ECE} \\
     \hline
     Logistic Regression & 3.61\% \\
     Random Forest &  4.39\% \\
     Support Vector Machine &   3.23\%\\
     Multi-Layer Perceptron &  3.59\%\\
     \hline
    \end{tabular}
\end{table}

The SVM model was identified as the best calibration-driven model, achieving a classwise-ECE of 3.23\% on the test set. This was followed by MLP, with a classwise-ECE of 3.59\%, and LR and RF models, with respective classwise-ECEs of 3.61\% and 4.39\% (see Table 2).

\begin{table}[h]
    \centering
    \caption{Accuracy achieved on the test set by each model along the accuracy branch}
    \begin{tabular}{ |p{4.5cm}|p{2.8cm}| }
     \hline
     \textbf{Model} & \textbf{Accuracy} \\
     \hline
     Logistic Regression & 65.69\% \\
     Random Forest &  65.34\% \\
     Support Vector Machine &  66.55\%\\
     Multi-Layer Perceptron &  65.69\%\\
     \hline
    \end{tabular}
\end{table}

The most accurate model was the SVM model with a test set accuracy of 66.55\%. The other candidate predictive models achieved accuracies ranging from 65.34\% to 65.69\% (see Table 3). As a result, SVM was identified as the best accuracy-driven model.

These two final models (calibration-driven SVM and accuracy-driven SVM) were used to generate predictions for the 2018/2019 NBA season. Taking as input these predictions along with the bookmaker's odds, we implemented and evaluated our betting systems. Each system was defined by a strategy and a rule. We tested one strategy (bet on all value bets) in combination with two different rules (fixed betting and Kelly betting), as described in section 7. Each simulation was carried out according to the algorithm described in Table 1.

\subsection{Fixed Betting}

The first rule we tested was the fixed betting rule. In this simulation, for each value bet identified by the model, the bettor placed a bet of \$100. We compare the performance of the calibration-driven and accuracy-driven betting systems in Fig. 3.

In this simulation, both systems identified several losing value bets early in the season (see Fig. 3 games 0-20). These losses were not significant, as under the fixed betting rule, the stake was restricted to \$100 dollars. The systems frequently identified value bets throughout the season, rarely opting not to bet on a game. While the pattern of betting appears similar between the two systems, the success enjoyed by each differed significantly. By the halfway-point of the season, the calibration-driven system was in the money, up approximately 25\% of its initial budget, while its accuracy-driven counterpart was down approximately 25\% (see Fig. 3 circa game 550). This suggests that there was a qualitative difference between the value bets identified by the models. The bets identified by the calibration-driven model were either successful more often, more profitable, or both, compared to those identified by the accuracy-driven model. This difference disappeared in the second half of the season, when the systems appeared to consistently place almost identical bets. As a result, the curves resemble mirror images of each other towards the end of the season, and both systems generated positive returns. However, the difference over the first half of the season proved decisive, and the gap between the bankrolls remained significant by season's end. The calibration-driven betting system ultimately achieved a highly profitable ROI of 32.45\%, while the accuracy-driven system achieved a respectable ROI of 5.56\%.

\begin{table}[h]
    \caption{Results of fixed betting simulations}
    \centering
    \begin{tabular}{ |p{4cm}|p{1.5cm}|p{1cm}| }
     \hline
     \textbf{Model} & \textbf{Final Bankroll}& \textbf{ROI} \\
     \hline
     Calibration-driven SVM &\$13,244.51 &32.45\% \\      
     Accuracy-driven SVM & \$10,556.29 & 5.56\%\\      
     \hline
    \end{tabular}    
    
\end{table}

\subsection{Kelly Betting}

Next, we tested a betting rule based on the Kelly criterion. In this simulation, for each bet identified by the model, the bettor calculated the stake as a function of the predicted probability and the odds offered. This was done using a conservative variation of the Kelly criterion known as the eighth-Kelly. Fig. 4 compares the performance of the calibration-driven and accuracy-driven betting systems under this rule.

Once again, both systems identified several losing value bets early in the season (see Fig. 4 games 0-20). This was followed by several spikes and dips for each system. Compared to the fixed betting systems, the bankrolls were much more volatile. This is because the loss on a single bet was capped at \$100 in the fixed betting systems, and capped at 12.5\% of the bankroll in the eighth-Kelly betting systems. This increase in volatility exacerbated the effect on the bankrolls of the qualitative difference in bets identified by each system in the first half of the season. As a result, by the time the accuracy-driven system's bankroll had approximately halved in size, the calibration-driven system's bankroll had increased by the same amount. (see Fig. 4 circa game 500). Volatility continued to be ever-present, and the calibration-driven system increased its bankroll by almost \$15,000 (to reach a peak of approximately \$20,000) over the span of approximately 100 games (see Fig. 4 circa game 800). While the models identified almost the exact same value bets over the second half of the season (as shown in Fig. 3), the curves look not at all similar towards the end of the season. In a Kelly-based betting system, simply identifying the same bets is not enough to achieve similar outcomes. In these systems, the size of the bankroll, as well as the predicted probability and the odds offered on each outcome, are crucial factors affecting the returns achieved. This is reflected in Fig. 4. The accuracy-driven system consistently diminished as the end of the season approached, ultimately achieving a negative ROI of -75.9\%. The calibration-driven system continued to experience high volatility, but mostly remained in the money, and concluded the season with an impressive ROI of 36.93\%.

\begin{table}[h]
    \caption{Results of Kelly betting simulations}
    \centering
    \begin{tabular}{ |p{4cm}|p{1.5cm}|p{1.25cm}| }
     \hline
     \textbf{Model} & \textbf{Final Bankroll}& \textbf{ROI} \\
     \hline
     Calibration-driven SVM &\$13,692.86 &36.93\% \\      
     Accuracy-driven SVM & \$2,409.66 & -75.9\%\\      
     \hline
    \end{tabular}    
\end{table}

\FloatBarrier
\begin{figure*}[t]
  \centering
    \includegraphics[width=\textwidth]{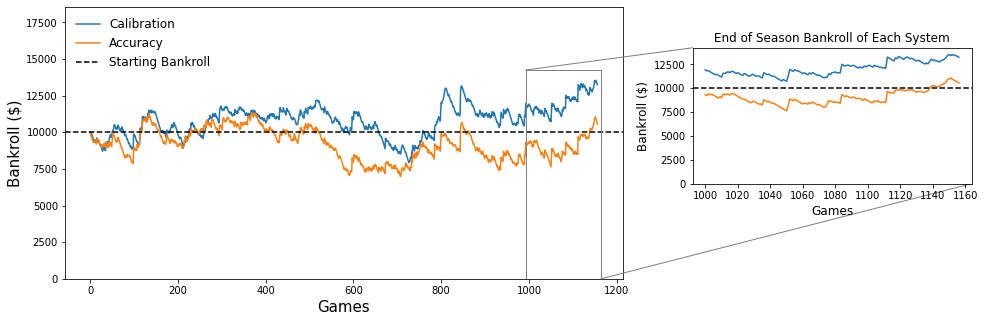}
    \caption{Bankrolls over the course of the 2018/2019 NBA season under the fixed betting rule. The blue curve represents the bankroll of the betting system using the calibration-driven predictive model, while the orange curve represents its accuracy-driven counterpart. The broken black line represents the initial bankroll. Curves which finish above this line earn a profit and are considered successful betting systems.}
\end{figure*}

\begin{figure*}[t]
  \centering
    \includegraphics[width=1\textwidth,left]{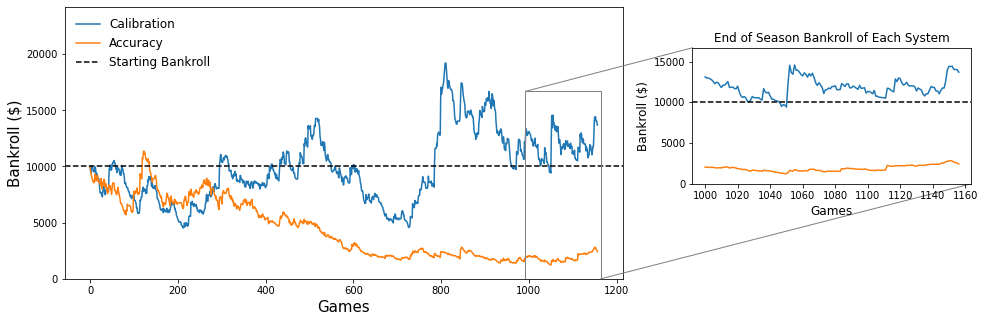}
    \caption{Bankrolls over the course of the 2018/2019 NBA season under the Eighth-Kelly betting rule. The blue curve represents the bankroll of the betting system using the calibration-driven predictive model, while the orange curve represents its accuracy-driven counterpart. The broken black line represents the initial bankroll. Curves which finish above this line earn a profit and are considered successful betting systems.}
    
\end{figure*}

\subsection{Evaluation of Central Hypothesis}

The experiment was designed to answer the question "does selecting a sports betting model based on calibration, rather than accuracy, allow for greater profit generation?". To answer this, we examine Table 6.

Both models frequently identified value bets, with calibration-driven systems betting on 87.55\% of games, and accuracy-driven systems betting on 89.89\% of games. Both models also had similar success rates, with the accuracy-driven systems winning 38.46\% of bets placed, and the calibration-driven systems winning a slightly higher 38.8\% of bets. Unsurprisingly, the accuracy-driven model was the more accurate of the two (accuracy of 64.62\% versus 64.27\%), while the calibration-driven model was more well-calibrated (classwise-ECE of 4.46\% versus 5.03\%). Ultimately, the metric of primary concern to bettors is ROI. Over the two rules (fixed betting and Kelly betting), the maximum ROI achieved by a calibration-driven betting system was 36.93\% (achieved using the eighth-Kelly betting rule to determine the stake). The maximum ROI achieved by an accuracy-driven system was 5.56\% (under the fixed betting rule).  Calibration-driven betting systems were also more profitable on average, with a highly lucrative average ROI of 34.69\%, compared to an ROI of -35.17\% as achieved by accuracy-driven betting systems on average.

\begin{table*}[t]
    \caption{Comparison of calibration-driven and accuracy-driven betting systems}
    \begin{tabular}{ |p{3.25cm}|p{2.5cm}|p{2.5cm}|p{1.75cm}|p{1.75cm}|p{1.5cm}|p{1.5cm}| }
     \hline
     \textbf{Model} & \textbf{\% Games Bet On}& \textbf{\% Bets Won}& \textbf{Accuracy}&\textbf{Classwise-ECE}&  \textbf{Maximum ROI}& \textbf{Average ROI} \\
     \hline
     Calibration-driven SVM &87.55\% &38.8\% &64.27\% &4.46\% & 36.93\% &34.69\%  \\
     Accuracy-driven SVM &89.89\% &38.46\% & 64.62\% & 5.03\% & 5.56\% &  -35.17\% \\
      \hline
    \end{tabular}    
\end{table*}

\smallskip
In the next section, we reflect on these results and discuss their implications.

\clearpage
\clearpage

\section{Discussion}

In this paper, we aimed to devise a data-driven approach to sports betting. 
Focusing on the NBA, we set out to show that it is possible to leverage data to make a profit over a single season. Identifying a gap in the literature, we hypothesised that accuracy is not the most appropriate metric to evaluate the performance of the predictive model in a sports betting system, and that betting systems would be more profitable if calibration was used instead.

To test this hypothesis, we designed two competing betting systems, one equipped with a predictive model selected based on calibration, the other with a model selected based on accuracy. Using these models to generate predictions for all games in an NBA season, we ran betting simulations where the systems identified value bets (betting opportunities where the predicted probability of victory was greater than the probability implied by the bookmaker's odds) and used either a fixed betting rule, or a variation of the Kelly criterion to determine the size of each bet. Measuring the returns achieved by each betting system, we were able to compare the profitability of calibration-driven systems and accuracy-driven systems. To the authors' knowledge, this work represents the first attempt to study the effect on profit generation of selecting a sports betting model on the basis of calibration, as opposed to the traditional approach of selecting the most accurate model.
Another novelty of this work comes in the form of the features used in the predictive model. While a common approach for NBA game outcome prediction is to use box score statistics averaged over the season to date, we show that averaging differences in box score statistics versus opponents over the season to date can result in similar success.

Basing model selection on calibration led to profitable betting systems in all cases, with an average ROI of  34.69\%, and an ROI of 36.93\% in the most profitable system. In contrast, basing model selection on accuracy led to an ROI of -35.17\% on average and 5.56\% in the best case, with the worst-performing system (which used the eighth-Kelly betting rule) losing over 75\% of its wealth. This reiterates the danger of using a Kelly-based betting system, especially when the model is not well-calibrated. In the case of our best-performing betting system, we showed that bettors can increase their wealth by more than a third over a single season. These exciting findings support our conjecture that in a data-driven sports betting system, basing model selection on calibration, rather than accuracy, leads to greater profit generation. This reiterates the findings of Hubá\v{c}ek and colleagues, who showed that optimising for decorrelation with the bookmaker's odds leads to greater returns than optimising for accuracy \cite{hubavcek2019exploiting}. 

Another interesting finding of our work relates to the difference in the games identified as value bets by each final model. The accuracy-driven SVM identified more value bets than the calibration-driven SVM (betting on 89.89\% of games compared to 87.55\%) and had a lower success rate (38.46\% versus 38.8\%). This could suggest a higher number of the value bets identified by the accuracy-driven model were false positives, i.e. betting opportunities where the odds were not actually favourable. This behaviour suggests the accuracy-driven model may have been overconfident in its predictions, compared to the calibration-driven model. This is a well known issue for traditional classification models \cite{kwok2000evidence}. It has been shown that in circumstances where models should be uncertain about the label (such as in regions of sparse data), the tendency is to output a more extreme, unrepresentative and overconfident prediction \cite{mackay1992evidence}. 
Examining the distribution of value bets identified by each model, we see evidence of this. Fig. 5 shows a scatter plot of the model's predicted probability versus the probability implied by the bookmaker's odds, for all value bets identified by the calibration-driven SVM, and the accuracy-driven SVM, respectively.

\begin{figure*}
  \centering
    \subfigure[Scatter plot of predicted probability versus implied probability for each game identified as a value bet by the calibration-driven SVM.]{\includegraphics[width=0.45\textwidth]{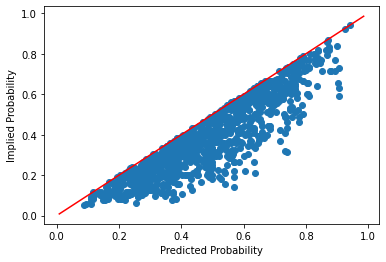}}
    \subfigure[Scatter plot of predicted probability versus implied probability for each game identified as a value bet by the accuracy-driven SVM.]{\includegraphics[width=0.45\textwidth]{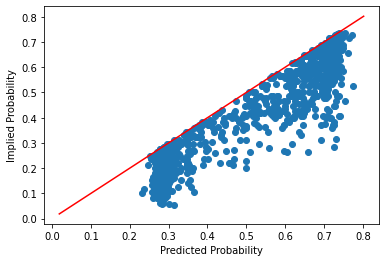}}    
    \caption{Comparison of scatter plots showing the predicted probability versus implied probability for each value bet identified by the predictive models (calibration-driven SVM, and accuracy-driven SVM, respectively).  All points lie below the line $y=x$ as a value bet is defined as one for which the model's predicted probability is greater than the probability implied by the bookmaker's odds ($x>y$).}
    \label{fig:my_figure}
\end{figure*}

We can see from Fig. 5 $(a)$ that the value bets identified by the calibration-driven model are approximately uniformly distributed (ignoring the extreme ends of the probability range). In contrast, the value bets identified by the accuracy-driven model are much more unevenly distributed (see Fig. 5 $(b)$). The bulk of these points lie in one of two regions, forming two clusters towards each end of the probability range, with much fewer value bets found in the middle region. Additionally, this set of predictions exhibits greater variance than its calibration-driven counterpart (0.036 versus 0.034). We can treat the implied probability as a proxy for the true probability \cite{wheatcroft2020profiting}. In this case, the accuracy-driven model's predictions appear further from the ground truth, compared to the calibration-driven model.
Knowing this, the fact that the accuracy-driven betting systems lost a higher percentage of bets comes as no surprise, as poorly calibrated models tend to be overconfident on incorrect predictions \cite{krishnan2020improving,guo2017calibration}.
This further emphasises the need for a well-calibrated model.

While the results are encouraging, a few critiques can be made. The most obvious is the use of a single season for the betting experiments, as no guarantee can be made that systems which were successful in this particular season would have been similarly profitable over other seasons. Further, the constraint imposed on predictions along the calibration branch of the predictive modelling pipeline (requiring at least 80\% of the bins to be non-empty) was somewhat arbitrary, and based on the authors' domain knowledge. Perhaps a better solution could be found to deal with the limitation of the classwise-ECE mentioned in section 3.

This research also leaves room for future work. One could experiment to find the  optimal betting strategy and rule to maximise profits. Another interesting idea would be to investigate the relationship between calibration and accuracy. Historically, bookmakers' accuracy in predicting NBA game winners is in the region of $69 \pm{} 2.5$\% \cite{hubavcek2019exploiting}. This begs the question, is there a limit to the accuracy that one can consistently achieve in predicting NBA game outcomes? If it exists, what is the limit that accuracy tends to, as the classwise-ECE tends to zero?

We have established a blueprint for developing a data-driven sports betting system, and shown that when evaluating predictive models for the sports betting problem, calibration is a more useful metric than accuracy. Beyond the realm of sports betting, calibration may be a more important metric than accuracy in any setting where the predicted probability is used for decision-making. This applies to many problems, such as weather forecasting and diagnosis of disease. Modellers who spend countless hours trying to increase the accuracy of probabilistic classifiers may be focusing on the wrong metric, and could be better served by trying to minimise the classwise-ECE. Practical applications of our results are clear - sports bettors can adopt our blueprint to increase their wealth. Finally, our findings can help  bookmakers too. Before setting the odds, the bookmaker generates their own predictions. They can use the classwise-ECE to reveal how far from the true probability their predictions lie. This could be an immensely valuable asset for their risk management team.

\section*{Acknowledgment}

This research builds upon work carried out as part of a dissertation for the degree of Master of Science in Data Science at the University of Bath. The authors would like to extend thanks to Dr. Alessio Guglielmi from the University of Bath for his support and guidance during this process. This research did not receive any specific grant from funding agencies in the public, commercial, or not-for-profit sectors. The authors would like to thank Sports Reference LLC for making NBA data available to researchers and basketball fanatics alike, and the American sportsbook Westgate for making their NBA odds data available on the website sportsbookreviewsonline.com. Finally, the authors are grateful to Prof. KongFatt Wong-Lin from Ulster University for his valuable comments and suggestions.



\bibliographystyle{elsarticle-harv} 
\bibliography{references}

\end{document}